\documentclass[10pt, conference, compsocconf]{IEEEtran}
\usepackage{cite}

\ifCLASSINFOpdf
  \usepackage[pdftex]{graphicx}
\else
\fi
\usepackage[cmex10]{amsmath}
\usepackage{algorithmic}

\usepackage{array}

\usepackage[tight,footnotesize]{subfigure}
\usepackage{url}

\usepackage{multirow}
\usepackage{kotex}
\usepackage{lipsum}

\begin{document}
%
\title{On the Ability of a CNN to Realize Image-to-Image Language Conversion}


\author{\IEEEauthorblockN{
Kohei Baba, 
Seiichi Uchida, 
Brian Kenji Iwana 
}
\IEEEauthorblockA{
Kyushu University, Fukuoka, Japan\\ Email: \{kohei.baba, brian\}@human.ait.kyushu-u.ac.jp
uchida@ait.kyushu-u.ac.jp}

}


%


\maketitle

\begin{abstract}
The purpose of this paper is to reveal the ability that Convolutional Neural Networks (CNN) have on the novel task of image-to-image language conversion. We propose a new network to tackle this task by converting images of Korean Hangul characters directly into images of the phonetic Latin character equivalent. The conversion rules between Hangul and the phonetic symbols are not explicitly provided. The results of the proposed network show that it is possible to perform image-to-image language conversion. Moreover, it shows that it can grasp the structural features of Hangul even from limited learning data. In addition, it introduces a new network to use when the input and output have significantly different features.
\end{abstract}


%
\IEEEpeerreviewmaketitle

\section{Introduction}
Deep neural networks, including convolutional neural networks (CNN) and recurrent neural networks (RNN), are well-known for their outstanding performance on various pattern recognition tasks.
Most of the state-of-the-art performances have been achieved by CNNs or RNNs, especially for document analysis and recognition (DAR) tasks. 
For example, scene text detection was considered to be a very difficult task about a decade ago but it now has become a rather tractable task by various CNN-based methods.
Cursive handwritten sentence recognition and historical document recognition are still difficult tasks but CNNs and RNNs are gradually increasing their state-of-the-art performance. 
\par
In addition to recognition tasks, CNN-based models, such as U-Nets~\cite{ronneberger2015u} and Generative Adversarial Networks~(GAN)~\cite{goodfellow2014generative}, are also used to image-to-image conversion tasks.
For example, the winner of the Document Image Binarization Competition 2017 (DIBCO2017)~\cite{pratikakis2017icdar2017} used a U-Net for successfully generating binarized images.
Recently, a stacked U-Net, called DocUNet~\cite{ma2018docunet} was applied to document flattening, or document unwarping, and has shown a good performance.
CNNs are also used for super-resolution for document images~\cite{datsenko2007example}.

\begin{figure}
	\begin{center}
	\small
        Input Image
    	\includegraphics[width=1.0\columnwidth]{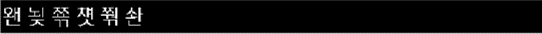}
        Output Image
        \subfigure[Image-to-Image Language Conversion]{\includegraphics[width=1.0\columnwidth]{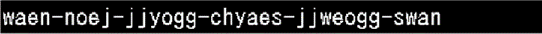}}
        \subfigure[The Proposed Semi-Convolutional Network]{\includegraphics[width=1.0\columnwidth,trim={.0cm .0cm .0cm 2.0cm},clip]{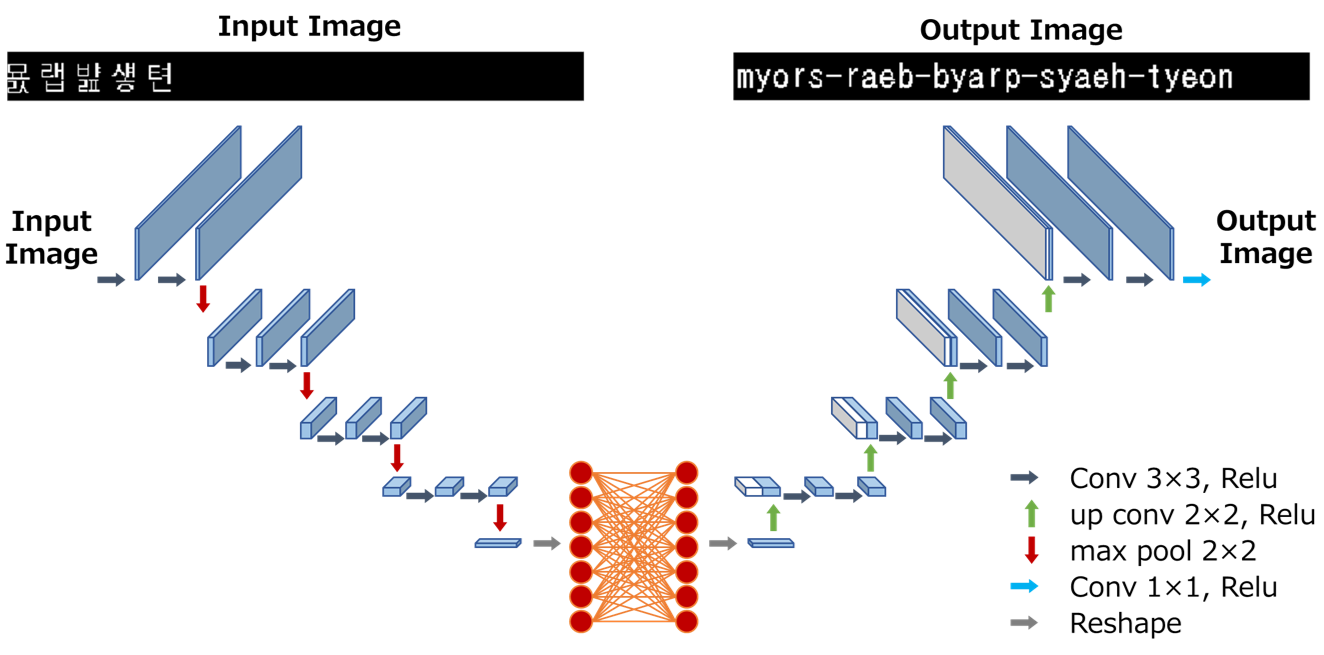}}
		\caption{(a)~Our image-to-image language conversion task. An {\em image} (i.e., a bitmap) of Korean Hangul characters is input and an {\em image} of the corresponding Latin phonetic symbols is output. (b)~The proposed network, called a Semi-Convolutional Network (SCN).  A fully-connected layer is introduced at the bottleneck part for information transmission across distant pixel locations.
		}
		\label{fig:network_architecture}
	\end{center}
	\vspace{-7mm}
\end{figure}

The purpose of this paper is to reveal the ability of CNNs on the {\it novel and benchmarking} task of image-to-image language conversion. Fig.~\ref{fig:network_architecture} demonstrates this task and the proposed CNN, called a Semi-Convolutional Network (SCN), used to tackle it.
Specifically, given a Korean Hangul character sequence {\it image}, a CNN outputs its phonetic Latin alphabet character sequence {\it image} counterpart.
In a microscopic view, the CNN has to perform a character-wise image conversion; each Hangul character is converted into 1 to 7 Latin characters.
In a macroscopic view, the CNN has to make large geometric conversions between images; an image containing $N$ Hangul characters is converted to an image containing $N$ to $7N$ Latin characters. This means the resulting Latin character sequence image is much longer than the original Hangul character sequence image. This also means that the output sequence has a variable length and the model must determine the spacing for each phonetic conversion.
%

		
Our trial has two main contributions.
First, we prove that our CNN can perform image-to-image language conversion very precisely.
It should be emphasized that this conversion is performed by CNN's information transmission ability across distant pixel locations.
In most image conversion tasks, such as denoising or binarization, a pixel in the original image is converted into the pixel at the same location in the destination image.
However, in our conversion task, the positions of corresponding characters are very distant, especially for the last characters of the Hangul character sequences. 
Specifically, due to the size difference between Hangul and the phonetic Latin equivalents, pixel information on the left side of the input image might relate to the right side of the output image. 
In order to address this, our CNN has creates a fully-connected mapping between the contracting path and expanding path in which pixels can affect other distant pixels. 
\par
The second contribution is that we prove that our CNN performs this conversion {\em without recognition of the characters}.
In other words, the CNN does not convert each individual character to its character code. In fact, during training, we do not give any character code sequence as ground truth nor do we predict any character code sequence as an output. 
Our CNN is trained purely with pairs of images, and no information is given except for image pairs.
To the authors' best knowledge, it is the first trial on this image-based character sequence conversion task.
Thus, this result will give a hint to other new trials on image-based character sequence conversion.
\par
The second contribution is crucial to realize that the proposed CNN can convert {\em unseen characters} to their correct Latin phonetic symbols. As we will see later, the CNN does not recognize individual Hangul characters but automatically learns the two-dimensional complex structure of each Hangul character which is comprised of several components representing vowels or consonants, and then determines the structure of the phonetic equivalent based only on visual information. 
%

Lastly, we have to explain the practical value of this paper. 
It would be easy to think that a character sequence in an image should be recognized into character codes for easier conversion. 
In fact, natural language processing tasks, such as translation, parsing, semantic analysis, and sentiment analysis, generally assume that their input is a sequence of character codes.
However, as stated as our second contribution, our method has the ability to convert unseen characters. This is impossible for code-based conversion methods; if a test character does not exist in the training set, it would have no reference to a code to be converted. 
In a sense, our method is real end-to-end learning without codes or rules and thus will be practical for the applications where the code system is not available.
\par
\section{Related Work}
\label{sec:related}
In general, using end-to-end learning to solve character conversion is more difficult than learning the independent sub-tasks of character recognition and then conversion~\cite{hoshen2016visual}. 
The sub-tasks themselves are well-researched areas. 
For example, there are many works on optical character recognition (OCR)~\cite{lecun1990handwritten,islam2017survey}. 
In recent years, neural networks have shown a high recognition accuracy for digital~\cite{uchida2016further}, natural scene~\cite{kwak2000video,Lee_2018,yi2013feature}, and handwritten characters~\cite{wan2013regularization,clanuwat2018deep}. 
Regarding Korean Hangul, methods of modeling its structural features have been proposed~\cite{kang2004utilization,simpson1994flexible} and recognition by neural networks have been used~\cite{cho1992recognition}.

This work also has similarities to other encoder-decoder networks such as Convolutional Encoder-Decoders~(SegNet)~\cite{Badrinarayanan_2017} and Deconvolution networks~(DeconvNet)~\cite{Noh_2015}. However, unlike U-Nets, SegNets, FCNs~\cite{Long_2015}, the proposed method is not fully convolutional as we propose the use of a full connection between the smallest convolutional layers. In addition, unlike many other encoder-decoder networks, there are no skip-connections or pooling index connections. Furthermore, the proposed SCN is more similar to a U-Net than a DeconvNet or SegNet in that there are no unpooling or the aforementioned pooling index connections for unpooling. Like U-Net, the upsampling is performed strictly through up-convolutions.

This work tackles language conversion which requires geometric conversions and large spatial transformations between the input and the output. 
This is unlike most image-to-image tasks which transform local regions of the input to the output. 
In a similar task, Hoshen et al.~\cite{hoshen2016visual} used a Multi-Layer Perceptron~(MLP) to perform image-to-image arithmetic demonstrating that some operations can be learned strictly from visual information. 

\section{Semi-Convolutional Network Architecture}
\label{sec:method}

We propose a new CNN-based model for image-to-image conversion tasks. The proposed network, shown in Fig.~\ref{fig:network_architecture}~(b), is called a Semi-Convolutional Network (SCN), as opposed to a Fully Convolutional Network (FCN)~\cite{Long_2015}. The architecture of the SCN is similar to a U-Net and consists of a contracting path (left side) and an expanding path (right side). The contracting path is similar to a CNN in that it has repeated applications of convolutional layers and max pooling layers. The expanding path also uses convolutional layers. However, instead of up-sampling between convolutional layer sets, SCN uses up-convolution layers to upscale the feature maps. 

The primary difference between a U-Net and an SCN is that the SCN is not fully convolutional. The traditional use of U-Nets is made of convolutional layers from end-to-end. 
However, the SCN replaces the convolutional layer at the transition between the contracting path and the expanding path with a full connection. 
The purpose of the full connection is to allow the network to transmit data across distant pixel locations. In a traditional U-Net, and FCNs in general, pixels outside of the largest receptive fields have no direct connection to each other. For the purpose of image-to-image language conversion, pixels of distant areas must be mapped to each other. 

In addition, compared to the traditional U-Net, the proposed SCN does not include skip-connections between the contracting and expanding layers. The purpose of skip-connections is to carry over fine details that would normally be lost in the contracting path~\cite{Long_2015}. This is done by copying the feature maps of the contracting path and concatenating them to the feature maps of the expanding path. However, for tasks where the input and output are not linearly related, they are not useful. We confirm this in the experiments. 
\section{Experimental Results}
\label{sec:experimental_results}
\subsection{Dataset}
\label{sec:dataset}
We observed the image-to-image language conversion ability of the SCN using Hangul, which is the modern Korean written language. 
Hangul characters consist of 19 consonant and 21 vowel components arranged in syllabic blocks~\cite{Taylor_1980}. Furthermore, the components have multiple arrangements depending on the sound. In total, there are 11,172 possible characters. As for the transliteration of Hangul into the Latin alphabet, we used the Romanization system implemented by Kromen~\cite{kromen}. 
Each Hangul character can be represented phonetically by 1 to 7 Latin characters. 

In the experiment, we created images with 1 to 10 horizontally aligned Hangul characters and their phonetic equivalent in Latin characters. 
All Korean and Latin characters are rendered by the fonts, ``SourceHanSans-Normal" and ``TakaoGothic," respectively. All images are 800 $\times$ 32 pixels regardless of the character sequence length. Therefore, for short sequences, the majority of the image is background. 

To confirm the advantage of image-to-image conversion, our training and test sets do not contain the same Hangul characters. Specifically, to construct the dataset, the 11,172 Hangul characters are divided into two character-independent pools, 8,937~($\sim$80\%) characters for the training set pool and 2,235~($\sim$20\%) characters for the test set pool. 
Then in order to simulate Hangul words, 1 to 10 Hangul characters are taken randomly from the respective character pools to create Hangul character sequences for the training and test sets. Due to the exponential number of possible combinations, only 89,370 and 22,350 sequences of Hangul characters and their Latin equivalent were generated, respectively. 
The dataset had an equal amount of 1 to 10 length Hangul character sequences. 

%

\subsection{Proposed and comparative evaluation details}

The following methods were used as comparative evaluations:

\paragraph{SCN (Proposed)}
The proposed method of using an SCN. In the implementation, the contracting path uses four sets of convolutional layers with two $3\times3$, stride 1 convolutional layers each. The sets are separated by $2\times2$ pooling with stride 2. The expanding path uses 2D up-sampling before each set of convolutional layers. The final layer is a convolutional layer with a $1\times1$ convolution. The number of nodes used in the convolutional layers is 8, 8, 16, 16, 32, 32, 64, and 64 for the contracting path and 64, 64, 32, 32, 16, 16, 8, 8, 1 for the expanding path. 
For the transition between the contracting path and the expanding path, a full connection is used between the flattened feature maps of the contracting path and the first layer of the expanding path.  
\paragraph{SCN w/ skip}
The same as the proposed SCN but with the same skip-connections proposed in the original U-Net implementation~\cite{ronneberger2015u}. This evaluation is to demonstrate the reason for the proposed method not using skip-connections.
\paragraph{MLP~\cite{hoshen2016visual}}
An MLP is used to demonstrate the problems with doing image-to-image language conversion with a standard neural network solution. The settings of the MLP were made to match a similar work~\cite{hoshen2016visual} which tackles image-to-image arithmetic. The MLP uses 3 hidden layers with 256 nodes each. The number of nodes in the output layer is equal to the number of pixels in the output. 
\paragraph{LSTM~\cite{Hochreiter_1997}}
Long Short-Term Memory~(LSTM) networks are RNNs which use gates to control the input, output, and forgetting of the internal states and are the standard for sequence-to-sequence modeling. 
In order to use an LSTM with the image-based character conversion dataset, we process the data by using columns of the images as elements of time steps. 
In this way, the height of the images relate to element dimensions and the width of the images is the time steps. 
The LSTM consisted of two layers.
\paragraph{E-D LSTM~\cite{Cho_2014}}
To evaluate the task by using a state-of-the-art sequence-to-sequence model, we adopt an Encoder-Decoder LSTM~(E-D LSTM). The difference between an E-D LSTM and the standard LSTM is that the E-D LSTM learns an embedding of the input to be decoded. The E-D LSTM uses two LSTM layers with 800 units and a fully-connected layer. 
\paragraph{U-Net~\cite{ronneberger2015u}} Since the proposed method is inspired by U-Net, it is the direct competitor. The hyperparameters used for this evaluation were set to match the SCN implementation with the exception of two convolutional layers with 64 nodes instead of the fully-connected layer in the SCN.

All of the networks are trained using cross entropy loss and an Adam optimizer with an initial learning rate of 0.001 for 100 epochs in batches of a size of 128. Also, the implementations of the convolutional and fully-connected layers use Rectified Linear Unit~(ReLU) activations with exception of sigmoid being used for the output layers.

\begin{table}[!t]
\renewcommand{\arraystretch}{1.1}
    \caption{Comparative results of the conversion of 1 to 10 length Hangul character sequences}
    \label{F_ED_overall}
    \centering
    \begin{tabular}{lccc}
    \hline
    Method & F-measure & Hamming distance & SED \\ \hline
    SCN (Proposed) &\textbf{0.988} &\textbf{5.09 $\times$ 10} &\textbf{0.86}\\ 
    SCN w/ skip&  0.987 & $5{.}40 \times 10$ & 1.18\\ \hline
    MLP~\cite{hoshen2016visual}   &0.717 &$ 1{.}07 \times10^3$ &24.5 \\
    LSTM~\cite{Schuster_1997}     &0.533 &$ 1{.}02 \times10^3$ &30.3 \\
    E-D LSTM~\cite{Cho_2014}      &0.846 &$ 6{.}39 \times10^2$ &16.3 \\
    U-Net~\cite{ronneberger2015u} &0.648 &$ 1{.}12 \times10^3$ &20.0 \\
    \hline
    \end{tabular}
	\vspace{-3mm}
\end{table}

\subsection{Evaluation metrics}
\label{sec:evaluation}
We used three measurements for the performance evaluation, F-measure, Hamming distance, and minimum string edit distance~(SED).
F-measure and Hamming distance are pixel-wise methods and the minimum SED evaluates the recognized character strings.

F-measure is the harmonic mean between the precision and the recall.
Hamming distance is the total number of incorrect pixels.
The F-measure and Hamming distance were calculated after binarizing the output image and the ground truth using a threshold of 0.5. 

Although F-measure and Hamming distance are reasonable for general image evaluations, they are too sensitive to our image conversion task. 
By missing one-pixel column or a single letter in the beginning (i.e., the left side) of the generated image, those measures are seriously degraded. Therefore, a more natural evaluation of this task would be to measure the accuracy of the generated characters. To accomplish this, we use the minimum SED between the recognized characters of the ground truth and the results. 

The SED is the minimum number of procedures to transform one string into another. A single procedure is defined as editing the string by insertion, deletion, or substitution of a character. In order to calculate the SED, we first use Tesseract OCR~\cite{smith2007overview}, an LSTM-based solution, to recognize the characters in the ground truth and the results images. Next, the recognized results are used with dynamic programming to find the minimum SED~\cite{Levenshtein}.

\subsection{Results}
\label{sec:results}

The results of 1 to 10 length Hangul character sequence conversion are shown in Table~\ref{F_ED_overall}. The proposed SCN had an F-measure of 0.988, Hamming distance of $5{.}09\times10$, and SED of 0.86, which is near perfect conversion. 

\begin{figure}
    \begin{flushright}
        \scriptsize
        Input: \raisebox{-.3\height}{\includegraphics[width=0.85\columnwidth,trim={2.2cm 1.62cm 1.8cm .3cm},clip]{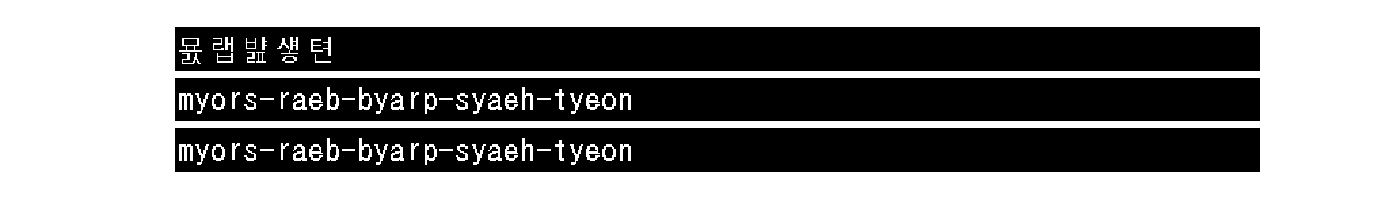}} \\
        \vspace{0.04cm}
        GT: \raisebox{-.3\height}{\includegraphics[width=0.85\columnwidth,trim={2.2cm .95cm 1.8cm .99cm},clip]{fig/output/vnet_without_skip/5.png}} \\
        \vspace{0.13cm}
        SCN: \raisebox{-.2\height}{\includegraphics[width=0.85\columnwidth,trim={2.2cm .95cm 1.8cm .99cm},clip]{fig/output/vnet_without_skip/5.png}} \\
        \vspace{0.04cm}
        U-Net: \raisebox{-.2\height}{\includegraphics[width=0.85\columnwidth,trim={2.2cm .95cm 1.8cm .99cm},clip]{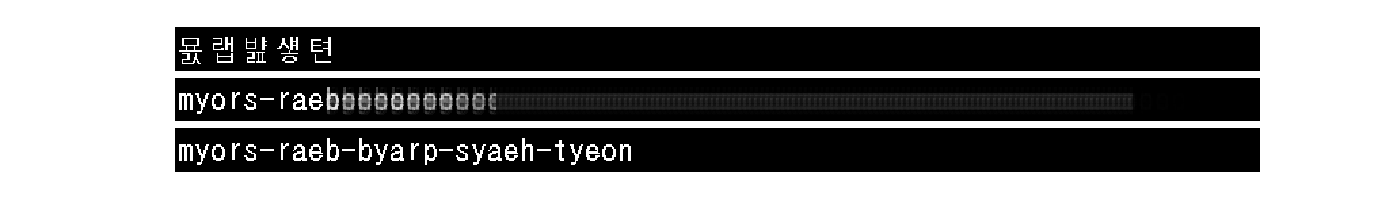}} \\
        \vspace{0.05cm}
        MLP: \raisebox{-.3\height}{\includegraphics[width=0.85\columnwidth,trim={2.2cm .95cm 1.8cm .99cm},clip]{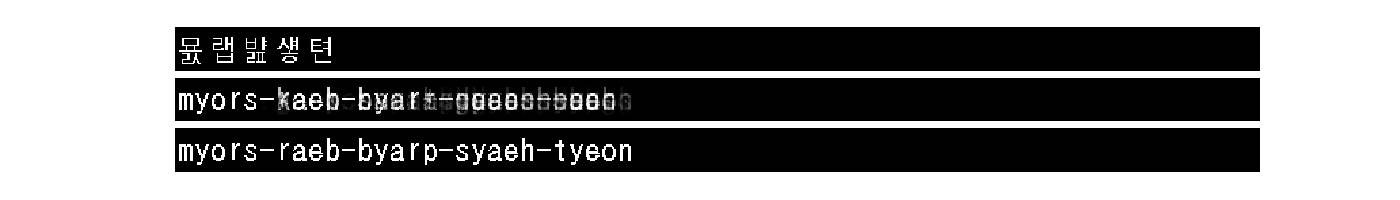}} \\
        \vspace{0.04cm}
        LSTM: \raisebox{-.3\height}{\includegraphics[width=0.85\columnwidth,trim={2.2cm .95cm 1.8cm .99cm},clip]{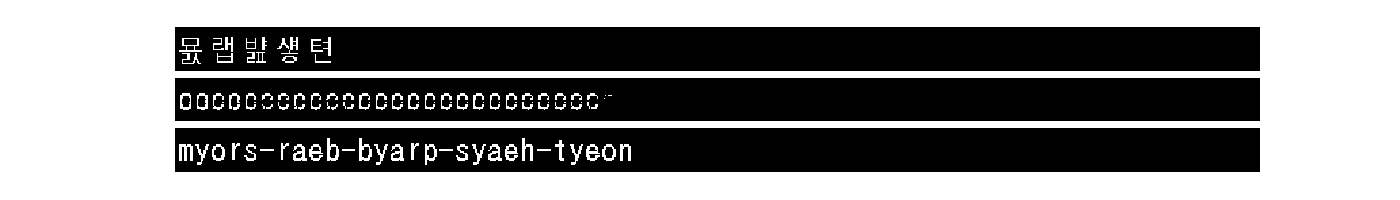}} \\
        \vspace{0.04cm}
        E-D~LSTM: \raisebox{-.3\height}{\includegraphics[width=0.85\columnwidth,trim={2.2cm .95cm 1.8cm .99cm},clip]{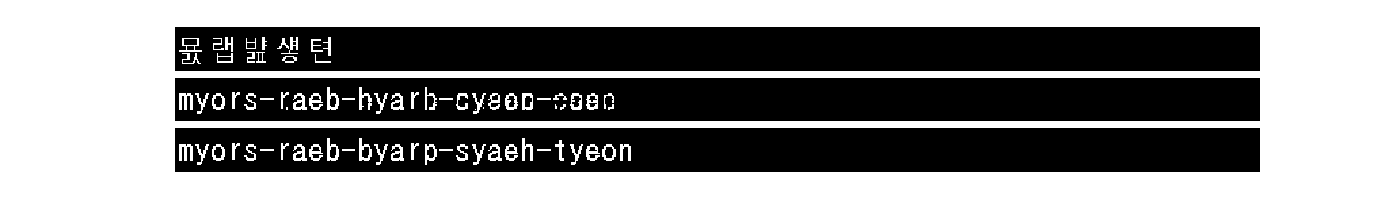}} \\
        \vspace{0.25cm}
        Input: \raisebox{-.3\height}{\includegraphics[width=0.85\columnwidth,trim={2.2cm 1.62cm 1.8cm .3cm},clip]{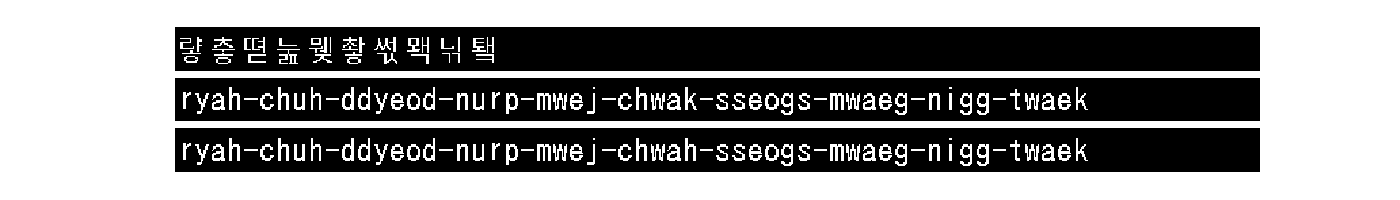}} \\
        \vspace{0.05cm}
        GT: \raisebox{-.3\height}{\includegraphics[width=0.85\columnwidth,trim={2.2cm .95cm 1.8cm .99cm},clip]{fig/output/vnet_without_skip/0.png}} \\
        \vspace{0.13cm}
        SCN: \raisebox{-.2\height}{\includegraphics[width=0.85\columnwidth,trim={2.2cm .95cm 1.8cm .99cm},clip]{fig/output/vnet_without_skip/0.png}} \\
        \vspace{0.05cm}
        U-Net: \raisebox{-.2\height}{\includegraphics[width=0.85\columnwidth,trim={2.2cm .95cm 1.8cm .99cm},clip]{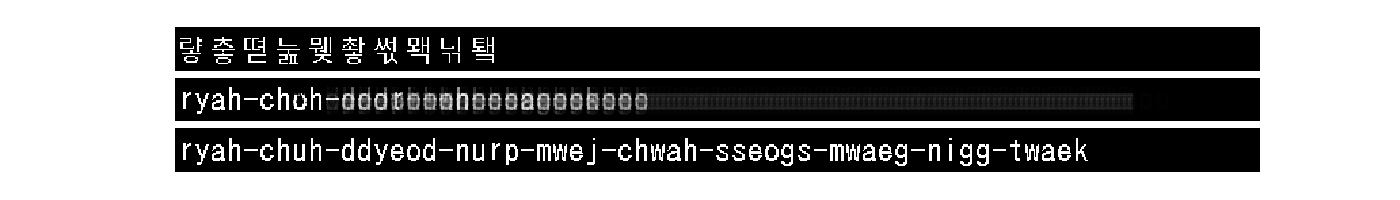}} \\
        \vspace{0.05cm}
        MLP: \raisebox{-.3\height}{\includegraphics[width=0.85\columnwidth,trim={2.2cm .95cm 1.8cm .99cm},clip]{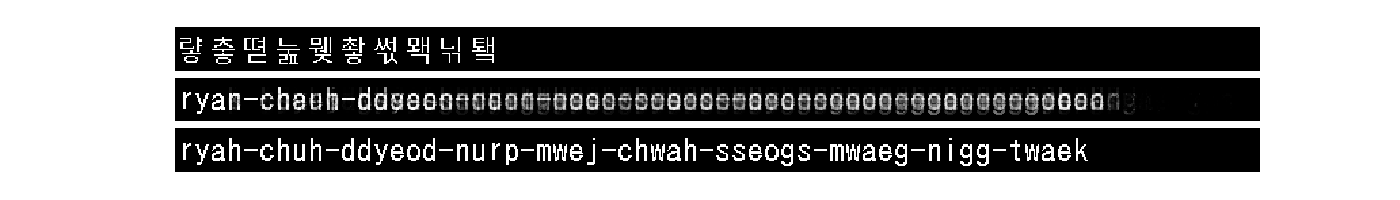}} \\
        \vspace{0.05cm}
        LSTM: \raisebox{-.3\height}{\includegraphics[width=0.85\columnwidth,trim={2.2cm .95cm 1.8cm .99cm},clip]{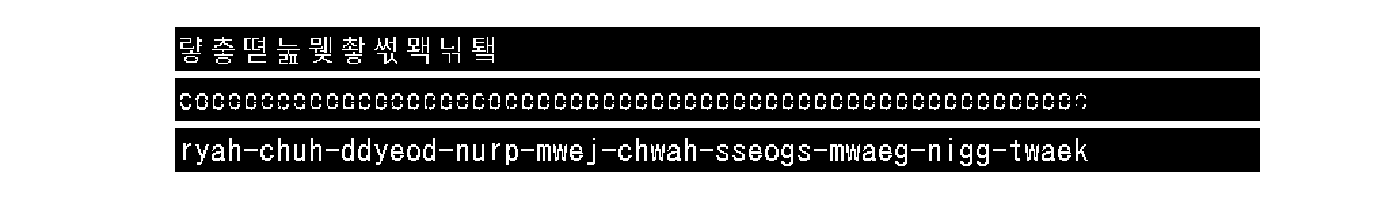}} \\
        \vspace{0.05cm}
        E-D~LSTM: \raisebox{-.3\height}{\includegraphics[width=0.85\columnwidth,trim={2.2cm .95cm 1.8cm .99cm},clip]{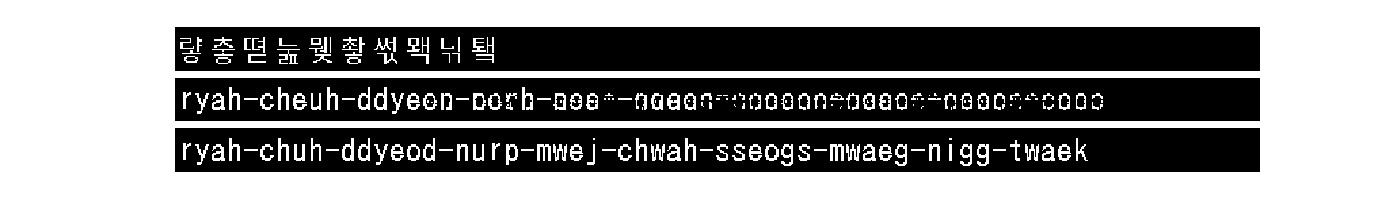}} 
	\end{flushright}\vspace{-4mm}
	\caption{Example results on image-to-image language conversion with 5 and 10 Hangul character sequence inputs. 
}
	\label{fig:example_results}
\end{figure}

Comparatively, all of the other methods had poor results. Fig.~\ref{fig:example_results} is an example of a 10 Hangul character sequence input. The U-Net had poor results due to the rigidity of being fully convolutional. 
Without the flexibility of the dense connections, input pixels of a U-Net can only affect output pixels within the receptive field of the smallest convolutional layer. The results in Fig.~\ref{fig:example_results} clearly demonstrates this with the first few Hangul characters converted correctly and the rest being noise. In addition, compared to all of the other methods, the U-Net results end abruptly due to limitations of the receptive field.

As for the MLP, characters early in the sequence were able to be successfully modeled. However, shown in Fig.~\ref{fig:example_results}, as the sequence progresses the MLP is unable to overcome the temporal problems associated with variable length character sequences. 

The LSTM and E-D LSTM also had poor results. The standard LSTM was used in a pure sequence-to-sequence fashion and this shows that it was unable to store detailed structured pixel information properly. However, it should be noted that the results of the plain LSTM were able to correctly estimate the length of the outputs, no matter how many characters were in the input. 

The E-D LSTM fared better than the LSTM due to being similar to the proposed method in that the input images are encoded and decoded from a fully-connected layer. In addition, the E-D LSTM performed the best out of the baselines. However, the results were worse than the proposed SCN. Like all of the models, E-D suffered the most when the input character sequences were long.   

%
\section{Analysis of Semi-Convolutional Network}
\label{sec:inspection}

\begin{figure}[t]
\small
\setlength\topsep{0pt}
    \begin{flushright}
		Input: \raisebox{-.2\height}{\includegraphics[width=0.88\columnwidth,trim={2.2cm 1.62cm 1.8cm .3cm},clip]{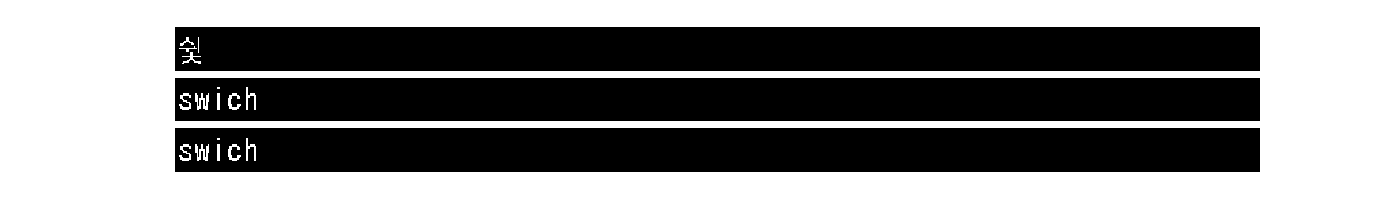}} \\
		SCN: \raisebox{-.3\height}{\includegraphics[width=0.88\columnwidth,trim={2.2cm .95cm 1.8cm .99cm},clip]{fig/good/0_F1_170.png}} \\
        \vspace{0.15cm}
		Input: \raisebox{-.2\height}{\includegraphics[width=0.88\columnwidth,trim={2.2cm 1.62cm 1.8cm .3cm},clip]{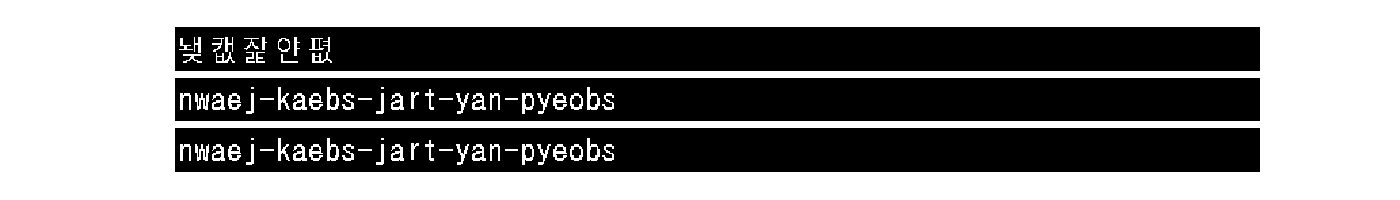}} \\
		SCN: \raisebox{-.3\height}{\includegraphics[width=0.88\columnwidth,trim={2.2cm .95cm 1.8cm .99cm},clip]{fig/good/4_F1_11168.png}} \\
        \vspace{0.15cm}
		Input: \raisebox{-.2\height}{\includegraphics[width=0.88\columnwidth,trim={2.2cm 1.62cm 1.8cm .3cm},clip]{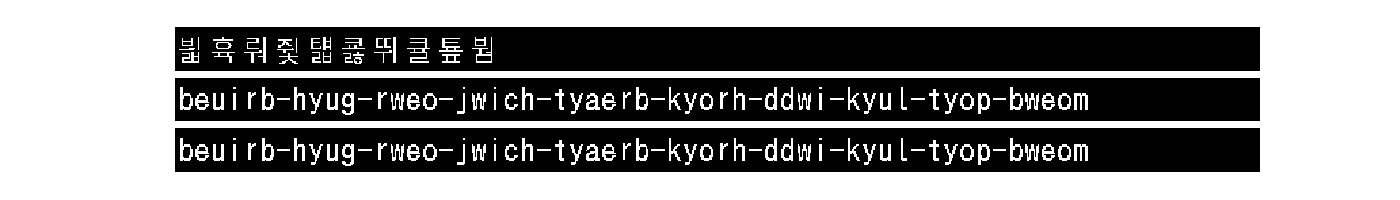}} \\
		SCN: \raisebox{-.3\height}{\includegraphics[width=0.88\columnwidth,trim={2.2cm .95cm 1.8cm .99cm},clip]{fig/good/9_F1_22345.png}} \\
	\end{flushright}\vspace{-2mm}
		\caption{Examples of SCN results with an F-measure of 1.0, meaning the result pixels are exactly the same as the ground truth. }
		\label{fig:good_results}
\end{figure}

\begin{figure}[t]
\small
\setlength\topsep{0pt}
        \hspace*{0.5cm} F-measure = 0.354\quad Hamming distance = $146$\quad \hspace{0.2cm}SED = 2
    \begin{flushright}
		Input: \raisebox{-.2\height}{\includegraphics[width=0.88\columnwidth,trim={2.2cm 1.62cm 1.8cm .3cm},clip]{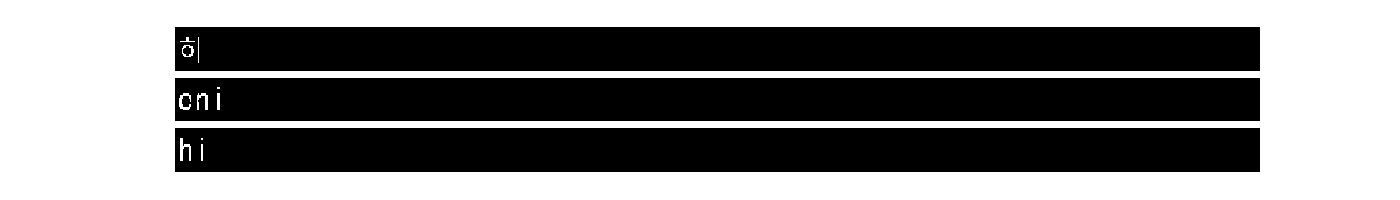}} \\
		GT: \raisebox{-.27\height}{\includegraphics[width=0.88\columnwidth,trim={2.2cm .27cm 1.8cm 1.55cm},clip]{fig/worst/0_F03539823008849557_724.png}} \\
		SCN: \raisebox{-.3\height}{\includegraphics[width=0.88\columnwidth,trim={2.2cm .95cm 1.8cm .99cm},clip]{fig/worst/0_F03539823008849557_724.png}} \\
	\end{flushright}
        \vspace{0.15cm}\hspace*{0.5cm} F-measure = 0.398\quad Hamming distance = $2{,}969$\quad SED = 9
    \begin{flushright}
		Input: \raisebox{-.2\height}{\includegraphics[width=0.88\columnwidth,trim={2.2cm 1.62cm 1.8cm .3cm},clip]{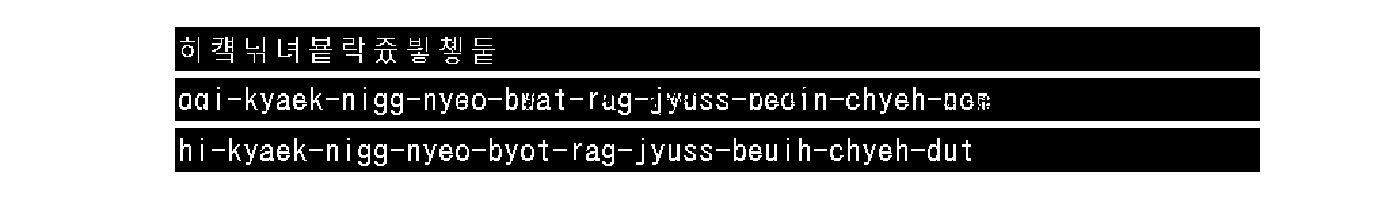}} \\
		GT: \raisebox{-.27\height}{\includegraphics[width=0.88\columnwidth,trim={2.2cm .27cm 1.8cm 1.55cm},clip]{fig/worst/9_F03983789260385005_21128.png}} \\
		SCN: \raisebox{-.3\height}{\includegraphics[width=0.88\columnwidth,trim={2.2cm .95cm 1.8cm .99cm},clip]{fig/worst/9_F03983789260385005_21128.png}} \\
	\end{flushright}
	    \vspace{0.15cm}\hspace*{0.5cm} F-measure = 0.408\quad Hamming distance = $1{,}584$\quad SED = 2
    \begin{flushright}
		Input: \raisebox{-.2\height}{\includegraphics[width=0.88\columnwidth,trim={2.2cm 1.62cm 1.8cm .3cm},clip]{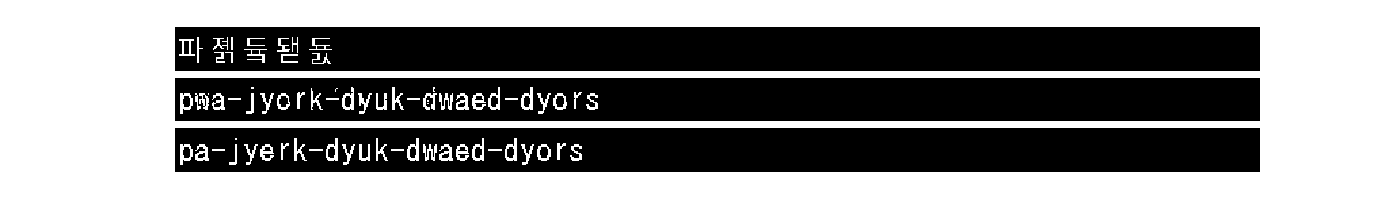}} \\
		GT: \raisebox{-.27\height}{\includegraphics[width=0.88\columnwidth,trim={2.2cm .27cm 1.8cm 1.55cm},clip]{fig/worst/4_F0408955223880597_8959.png}} \\
		SCN: \raisebox{-.3\height}{\includegraphics[width=0.88\columnwidth,trim={2.2cm .95cm 1.8cm .99cm},clip]{fig/worst/4_F0408955223880597_8959.png}} \\
	\end{flushright}\vspace{-2mm}
		\caption{The SCN results with the three lowest F-measures. }
		\label{fig:bad_results}
		\vspace{-2mm}
\end{figure}

The proposed SCN had much better results compared to any other method. Fig.~\ref{fig:good_results} shows example results from SCN with perfect F-measures using characters sequences of different lengths. The SCN is able to do image-to-image language conversion correctly, even for times when the length of Hangul characters in the input was long. 

\begin{figure}[t]
\small
\setlength\topsep{0pt}
    %
    \hspace*{0.5cm} F-measure = 0.767\quad Hamming distance = $1{,}594$\quad SED = 34\par
    \begin{flushright}
		Input: \raisebox{-.2\height}{\includegraphics[width=0.88\columnwidth,trim={2.2cm 1.62cm 1.8cm .3cm},clip]{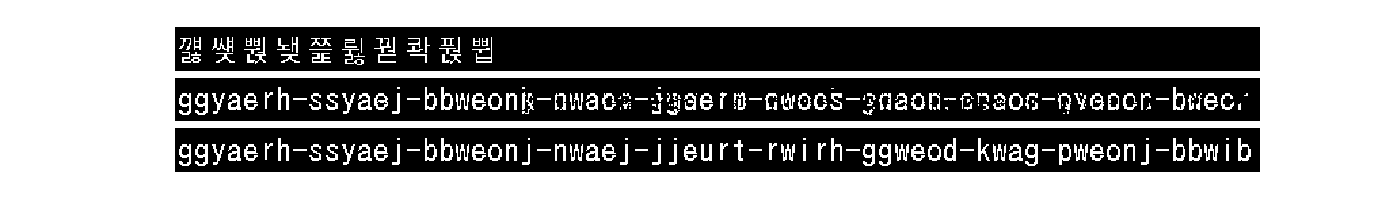}} \\
		GT: \raisebox{-.27\height}{\includegraphics[width=0.88\columnwidth,trim={2.2cm .27cm 1.8cm 1.55cm},clip]{fig/worst/9_SED350_21690.png}} \\
	\end{flushright}
	\hspace*{0.065cm} OCR:
    {\fontsize{6.5pt}{0pt}\selectfont ggyaerh-ssyaej-bbweonj-nwaej-jjeurt-rwirh-ggweod-kwag-pweonj-bbwib \par}
    \begin{flushright}
		SCN: \raisebox{-.3\height}{\includegraphics[width=0.88\columnwidth,trim={2.2cm .95cm 1.8cm .99cm},clip]{fig/worst/9_SED350_21690.png}} \\
	\end{flushright}
	\hspace*{0.065cm} OCR:
    {\fontsize{6.5pt}{0pt}\selectfont ggyaerh-ssyaej-bbweoni-nwao:2-:j:§aerm-nwaes-grzaonrmam-manor:-bu:ec.‘ \par}
    %
    \vspace{0.15cm}\hspace*{0.5cm} F-measure = 0.735\quad Hamming distance = $1{,}381$\quad SED = 32
    \begin{flushright}
		Input: \raisebox{-.2\height}{\includegraphics[width=0.88\columnwidth,trim={2.2cm 1.62cm 1.8cm .3cm},clip]{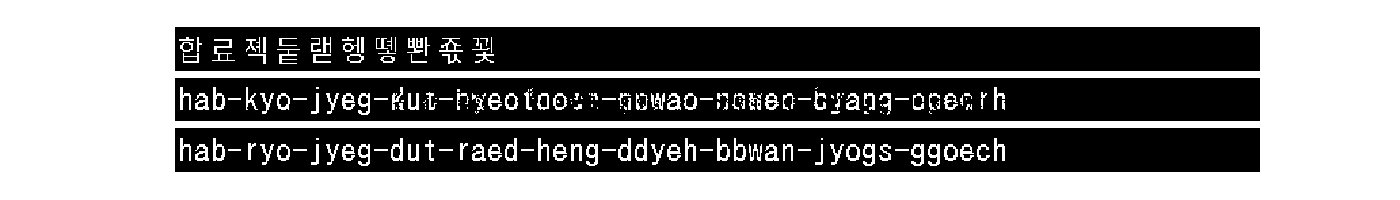}} \\
		GT: \raisebox{-.27\height}{\includegraphics[width=0.88\columnwidth,trim={2.2cm .27cm 1.8cm 1.55cm},clip]{fig/worst/9_SED310_21092.png}} \\
	\end{flushright}
	\hspace*{0.065cm} OCR:
    {\fontsize{6.5pt}{0pt}\selectfont hab-ryo-jyeg-dut-raed-heng-ddyeh-bbwan-jyogs-ggoech \par}
    \begin{flushright}
		SCN: \raisebox{-.3\height}{\includegraphics[width=0.88\columnwidth,trim={2.2cm .95cm 1.8cm .99cm},clip]{fig/worst/9_SED310_21092.png}} \\
	\end{flushright}
	\hspace*{0.065cm} OCR:
	{\fontsize{6.5pt}{0pt}\selectfont hab-kyo-jyeg-fiui-b’g’eot‘no:-r-nuwao-maen-hg‘agg-opecrh \par}
    %
	\vspace{0.15cm}\hspace*{0.5cm} F-measure = 0.670\quad Hamming distance = $1{,}791$\quad SED = 31
    \begin{flushright}
		Input: \raisebox{-.2\height}{\includegraphics[width=0.88\columnwidth,trim={2.2cm 1.62cm 1.8cm .3cm},clip]{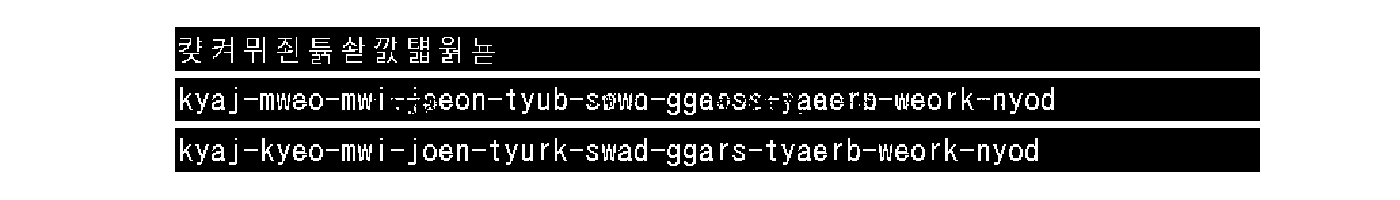}} \\
		GT: \raisebox{-.27\height}{\includegraphics[width=0.88\columnwidth,trim={2.2cm .27cm 1.8cm 1.55cm},clip]{fig/worst/9_SED300_20129.png}} \\
    	\end{flushright}
    	\hspace*{0.065cm} OCR:
        {\fontsize{6.5pt}{0pt}\selectfont kyaj-kyeo-mwi-joen-tyurk-swad-ggars-tyaerb-weork-nyod \par}
        \begin{flushright}
		SCN: \raisebox{-.3\height}{\includegraphics[width=0.88\columnwidth,trim={2.2cm .95cm 1.8cm .99cm},clip]{fig/worst/9_SED300_20129.png}} \\
	\end{flushright}
	\hspace*{0.065cm} OCR:
	{\fontsize{6.5pt}{0pt}\selectfont kyaj-mweo-mwi'rfiqsof‘i-‘lyub-sfiwo-qqbbs‘er3-‘aeera-weork-nyod \par}

		\caption{The SCN results with the three highest SEDs. The character sequences under the GT and the output image are OCR results.}
		\label{fig:bad_results2}
		\vspace{-4mm}
\end{figure}

Figs.~\ref{fig:bad_results} and~\ref{fig:bad_results2} show examples where the SCN could not model the results perfectly. In Fig.~\ref{fig:bad_results}, the SCN result images with the worst three F-measures are shown. The common factor between them is that the first Hangul character has a short phonetic equivalent. Specifically, ``히" and ``파" are converted to two character equivalents, ``hi" and ``pa," respectively. The problem is that one and two character equivalents only make up 2.9\% of the total possible conversions. Thus, the SCN attempted to convert the leading character into three character conversions, which caused for cascading failure of the F-measure and Hamming distance.
As for Fig.~\ref{fig:bad_results2}, the images with Hangul characters that require many abnormally long Latin equivalents caused large SEDs for the SCN. 

\subsection{On skip-connections}
One important difference between U-Nets and the proposed SCN is the lack of skip-connections. In order to observe the effect the skip-connections has on image-to-image language conversion, we performed an experiment with the skip-connections intact. Table~\ref{F_ED_overall} shows that the skip-connections are detrimental to the overall performance of the SCN. 
This degradation is justified due to the skip-connections intentionally carrying structural information from the contracting layers which are counterproductive to the task. 

\subsection{Individual character length evaluations}

\begin{figure}[!t]
\centering
\subfigure[F-Measure]{\includegraphics[width=0.49\columnwidth]{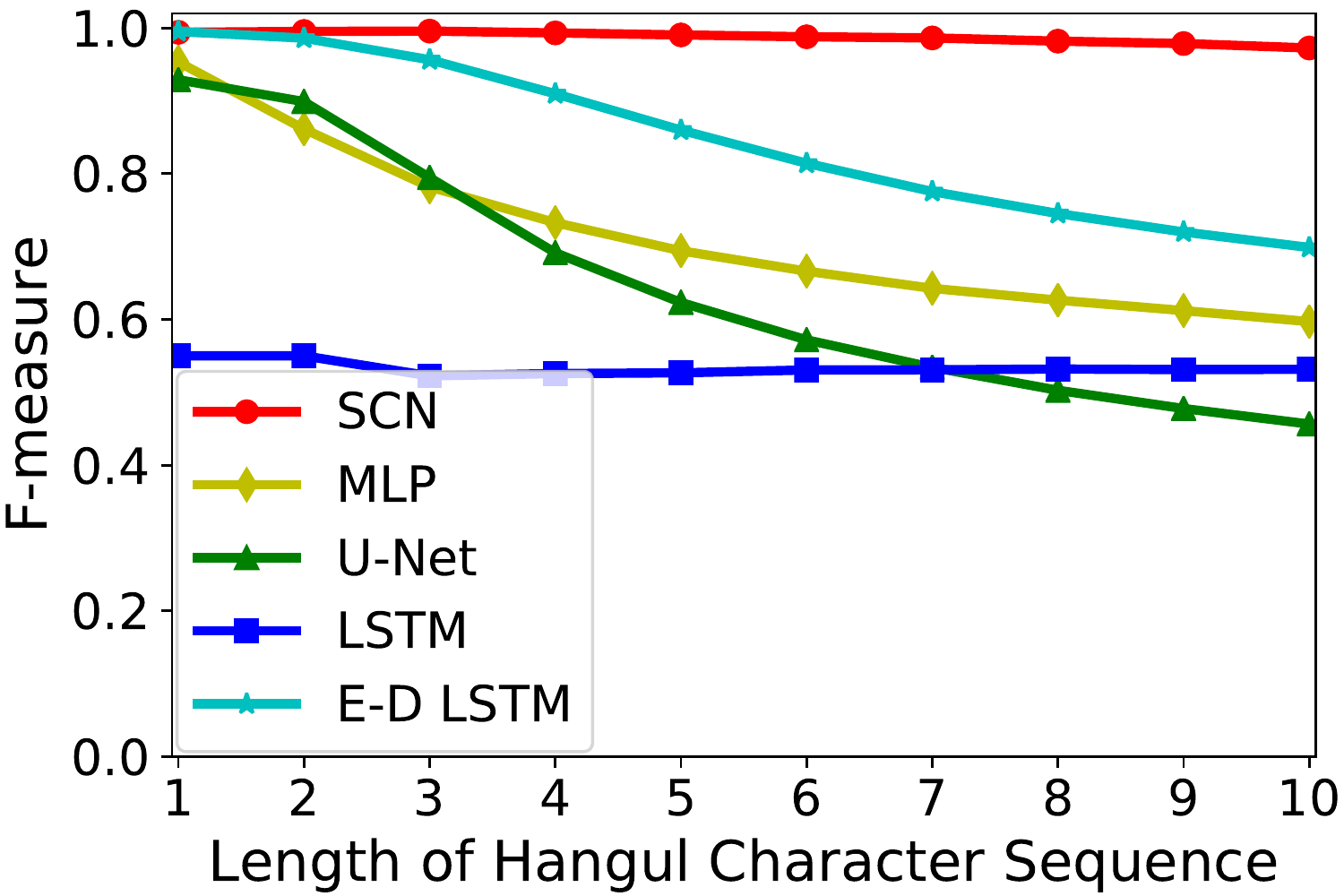}}
\subfigure[SED]{\includegraphics[width=0.49\columnwidth]{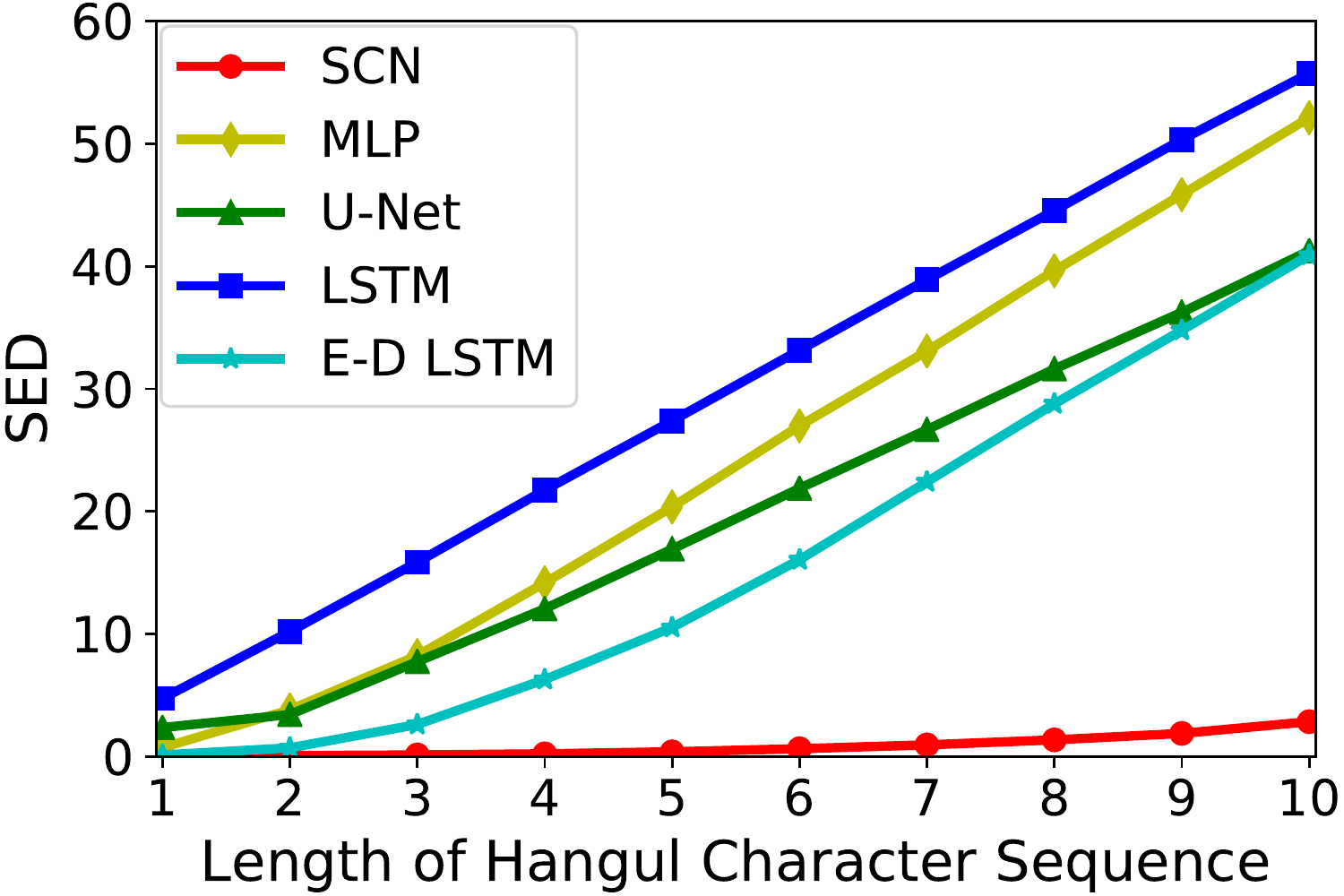}}
\caption{The average (a) F-measure and (b) SED of the evaluated methods at different length of Hangul characters. The different Hangul character sequence lengths were trained in one model and are only evaluated separately.}
\label{fig:number}
\end{figure}

\begin{figure}[!t]
\centering
\subfigure[F-Measure]{\includegraphics[width=0.49\columnwidth]{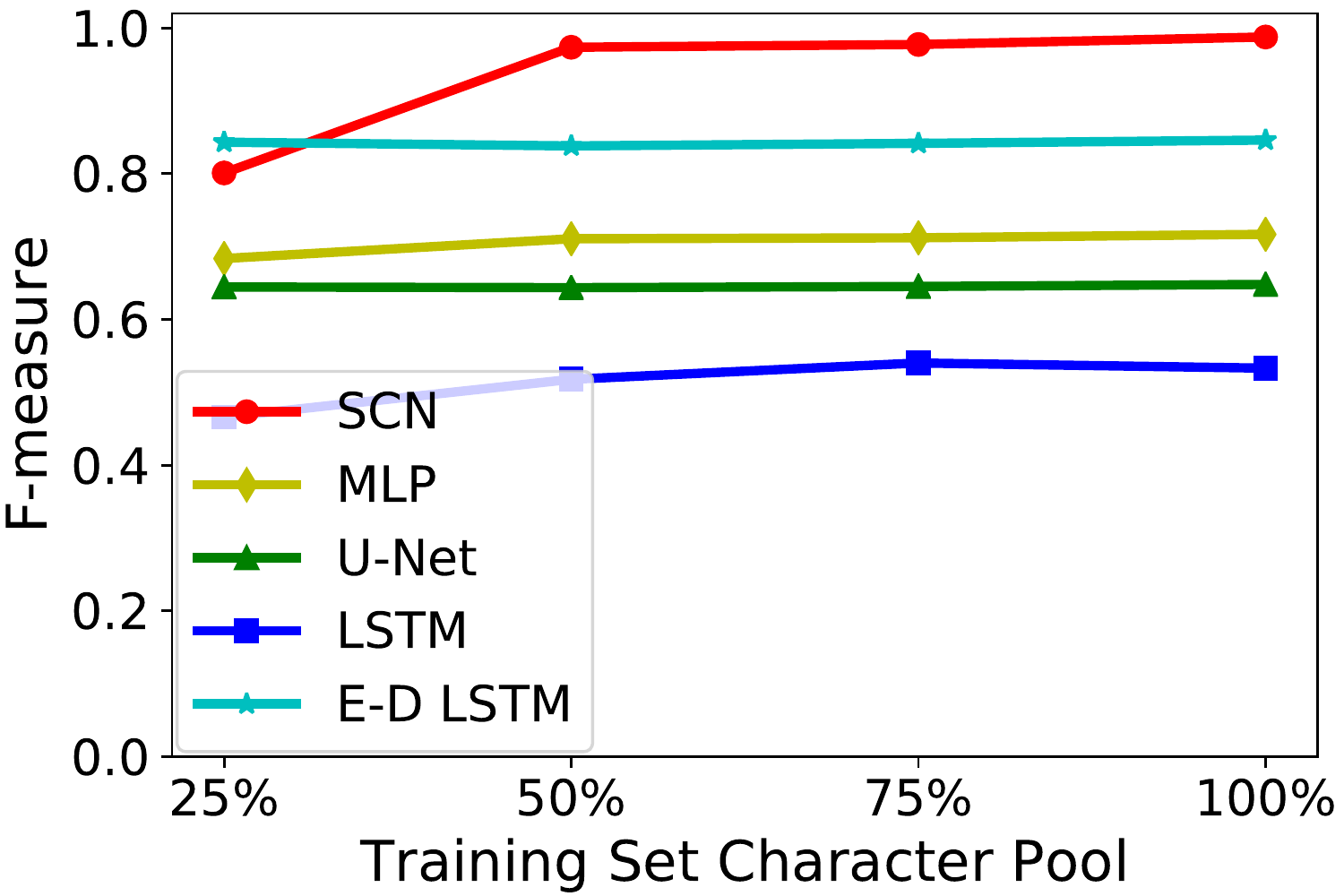}}
\subfigure[SED]{\includegraphics[width=0.49\columnwidth]{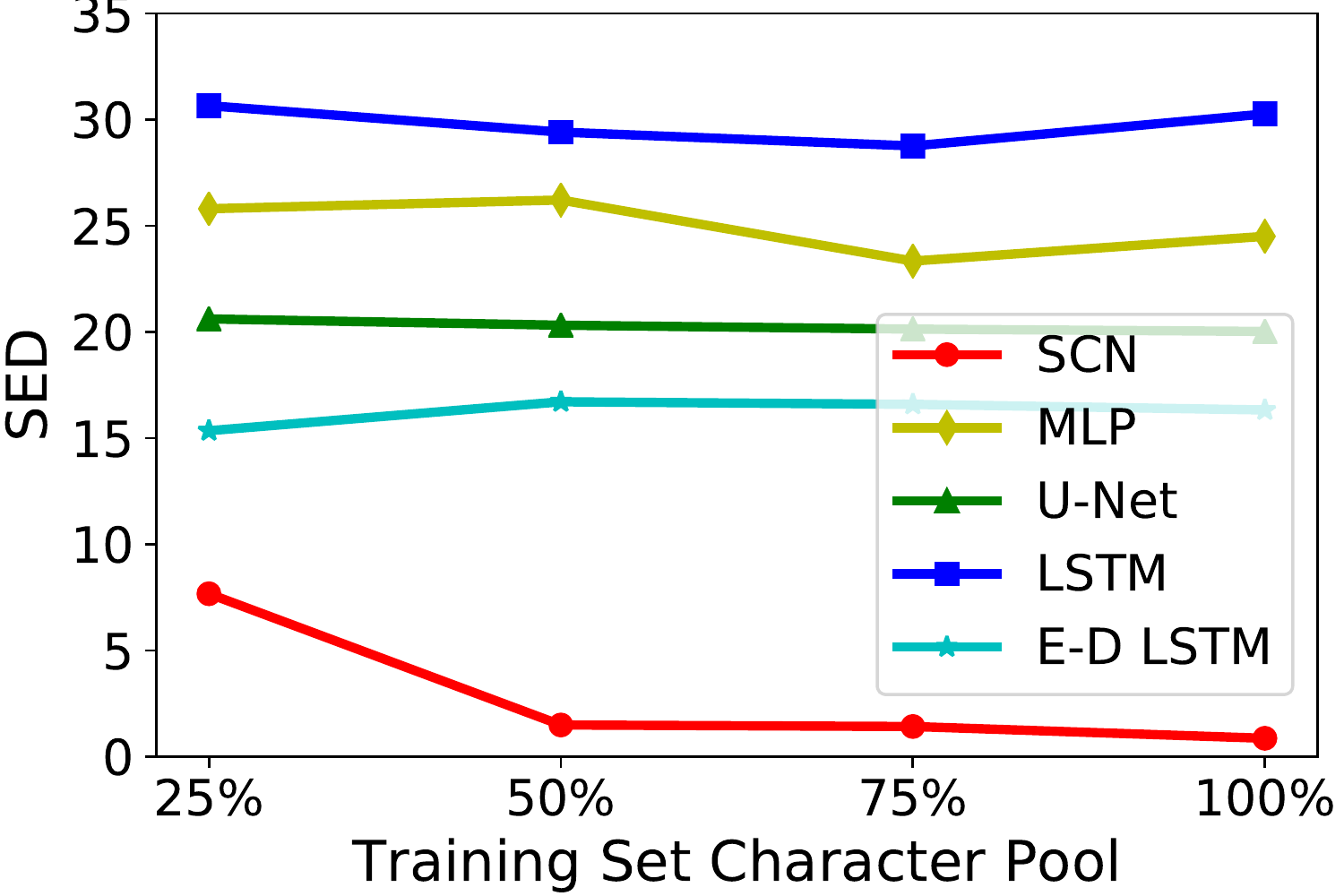}}
\caption{The average (a) F-measure and (b) SED of the evaluated methods at different percentages of training set character pool.}
\label{fig:ratio}
\vspace{-4mm}
\end{figure}

In order to understand the reason for SCN's success, we evaluated the lengths of the Hangul sequences separately. Fig.~\ref{fig:number} shows that the SCN is able to maintain a high accuracy for every input Hangul sequence length.
For the other evaluated methods, there is a rapid decrease in F-measure and an increase in SED as the number of characters in the sequences grows. However, in one instance, E-D LSTM had a slightly higher average F-measure than SCN. For 1-Hangul character inputs, E-D LSTM has an average F-measure of 0.9951 and the proposed method at 0.9940. On the other hand, for all other input characters lengths, E-D LSTM does significantly worse. 

\subsection{Effect of training set size}
A typical character or word recognition method can only convert entries within its corpus. 
However, since the spatial components of the images can be used, image-to-image language conversion allows the conversion of unseen characters.
The proposed method demonstrates this ability by using character-independent training and test sets. 

In order to test the limit on what can be converted, additional experiments were performed using 25\%, 50\%, and 75\% of the training set character pools. ``25\%" means that 25\% of the 8,937 possible training characters are used to generate the training set. In other words, for ``25\%," only 2,235 possible Hangul characters are used to generate the 89,370 Hangul character sequence images for the training set. 
The test sets were fixed to be the same as in Section IV. 
Fig. ~\ref{fig:ratio} is the results of using the limited training sets. For the proposed method, the F-measure does not dramatically decrease until ``25\%." This shows that even with limited training characters, the SCN can produce the phonetic equivalent for many unseen Hangul characters.

\section{Conclusion}
\label{sec:conclusion}
In this paper, we proposed a new image-to-image CNN called a Semi-Convolutional Network (SCN), which is similar to the popular U-Net. 
The difference is that an SCN contains a fully-connected layer at the transition from the contracting path and the expanding path. This gives it the advantage of being able to be used as a solution for tasks that require information to be transferred between distant pixel locations.

In order to demonstrate its ability, we explore the novel task of image-to-image language conversion. 
The proposed SCN is tasked with learning image-to-image Korean Hangul to phonetic Latin language conversion without explicitly defining the conversion rules.
Consequently, the proposed network allows for the conversion of unseen characters which is impossible with any dictionary or code-based methods. 
We conducted experiments on images of simulated Korean words using 1 to 10 length Hangul character sequences. 
The results showed that perfect conversion is possible in many cases regardless of the length of input characters in Hangul.
Furthermore, the proposed SCN had significantly better performance than every comparison method.

This work is the first step in realizing end-to-end language conversion and it demonstrates the ability to use a neural network to perform an image-to-image task which would be traditionally done in sub-tasks. 
Future work could extend SCNs to many new applications that U-Net is not suited for due to the receptive field constraints.

\section*{Acknowledgement}
This work was supported by JSPS KAKENHI Grant Number JP17H06100.







\newcommand{\BIBdecl}{\setlength{\itemsep}{0.980mm}}

\bibliographystyle{IEEEtran}
\bibliography{translation}
%



\end{document}